\documentclass[runningheads]{llncs}
\usepackage{graphicx}
\usepackage{booktabs}
\usepackage{amsmath}
\usepackage{graphicx}
\usepackage{hyperref}
\usepackage[accsupp]{axessibility}
\usepackage{marvosym}
\usepackage{orcidlink}
\makeatletter
\def\blfootnote{\gdef\@thefnmark{}\@footnotetext}
\makeatother

\begin{document}
\title{LARAD: Layout-Aware Road Anomaly Detection via Spatial-Logic Reasoning}
\titlerunning{Layout-Aware Road Anomaly Detection}
%
\author{Shiyi Mu\inst{1} \and
Xujie Chen\inst{1} \and \\ 
Shugong Xu\textsuperscript{\Letter}\inst{2}}
\authorrunning{S. Mu et al.}
%
\institute{Shanghai University, Shanghai, China \\ \email{shiyimu@shu.edu.cn} \and
Xi’an Jiaotong-Liverpool University, Suzhou, China\\
\email{shugong.xu@xjtlu.edu.cn}}

\maketitle              

\blfootnote{\textsuperscript{\Letter} Shugong Xu is the  corresponding author.}

\begin{abstract}

Accurate open-world obstacle detection is critical for autonomous driving. Current anomaly segmentation methods suffer from a fundamental blind spot: they over-rely on texture novelty to identify out-of-distribution (OoD) objects while ignoring contextual spatial logic. Furthermore, mitigating the resulting false positives often requires cascading massive vision models, introducing unacceptable inference latency. To address these issues, we propose Layout-Aware Road Anomaly Detection (LARAD), shifting the paradigm from appearance matching to spatial-logic reasoning. First, we introduce the Spatial-Logic Violation Synthesis (SLVS) pipeline, which generates training samples that are texture-consistent yet spatially invalid, forcing the model to learn contextual violations. Second, we augment a standard closed-set segmentation network with a lightweight, OoD-guided attention branch. Extensive experiments demonstrate that LARAD significantly enhances robustness against logical anomalies and establishes a new state-of-the-art, all while retaining the high efficiency of a single-model architecture.

\end{abstract}
\section{Introduction}
\label{sec:intro}

Segmentation algorithms have evolved from closed-set tasks, such as semantic and panoptic segmentation, to highly flexible open-world paradigms driven by prompt-based models and free-form textual descriptions. As autonomous driving and robotic perception systems transition from constrained, predictable environments to the complex open world, road segmentation faces severe open-set challenges. These challenges stem primarily from the critical need to identify novel, unknown obstacles of arbitrary categories to ensure driving safety\cite{S3AD}\cite{DDStereo}. Effectively addressing this open-world obstacle detection requires overcoming two fundamental bottlenecks: the data synthesis paradigm and the algorithmic design.

From a data perspective, early open-set segmentation relied solely on anomaly-free images, as exhaustive collection of real-world anomaly scenarios is inherently intractable and prohibitively expensive. To bridge this gap, later works introduced synthetic anomalies via COCO-based paste-in techniques or diffusion models, leveraging out-of-distribution (OoD) samples to separate known and unknown feature spaces. However, these paradigms suffer from a critical flaw: texture bias. By predominantly introducing novel categories from external domains, they inadvertently train models to detect mere appearance shifts (unusual textures) rather than genuine contextual anomalies. Consequently, these models struggle with spatial-logic anomalies objects that naturally exist in the environment like traffic cones and trash bins, but pose severe safety risks when placed on the driving lane. To break this texture reliance, we redefine the anomaly synthesis pipeline. We introduce Spatial-Logic Violation Synthesis (SLVS), which constructs texture-consistent yet spatially-invalid training samples by repositioning indigenous background objects into road regions, forcing the model to learn contextual layout violations rather than shortcutting via texture novelty.

Algorithmically, early approaches for road anomaly segmentation relied heavily on reconstruction inconsistencies or prediction uncertainties, such as Maximum Softmax Probability (MSP) and Mask2Former\cite{Mask2Former} based confidence analysis as M2A~\cite{Mask2Anomaly} and RbA~\cite{RbA}. To suppress false positives caused by background over-detection, recent state-of-the-art methods rely on cascaded refinement pipelines. For instance, S2M~\cite{S2M_CVPR24} and OoDDINO~\cite{OoDDINO} convert pixel-level anomaly scores into 2D bounding boxes and feed them into massive foundation models like SAM\cite{SAM} for post-processing. While these multi-model cascaded paradigms effectively filter background noise, they incur severe computational overhead and latency, making them highly impractical for real-time deployment on resource-constrained autonomous vehicles.

In contrast, rather than relying on an external, computationally heavy detection stage, we propose a cascade-free, layout-aware solution. We augment the efficient EoMT~\cite{EoMT} segmentation network with a lightweight OoD-guided attention layer at its output stage. Crucially, this mechanism incorporates local spatial relationship modeling, enabling the network to perceive the contextual layout of the scene. By explicitly restricting dedicated, learnable OoD queries to focus solely on high-uncertainty regions guided by OoD heatmap, the model directly refines these heatmap into precise anomaly masks. Benefiting from this streamlined architecture, our approach entirely avoids the latency bottleneck of cascaded models, retaining the inference efficiency of a single-model solution without compromising the original closed-set segmentation performance.

\noindent Our main contributions are:
\begin{itemize}
    \item We propose a cascade-free, single-model architecture that augments standard segmentation networks with spatial-logic reasoning, ensuring highly efficient and precise road anomaly localization.
    \item We design a lightweight module that restricts dedicated OoD queries to high-uncertainty regions and encodes local context, effectively eliminating the latency of multi-stage post-processing.
    \item We introduce a novel synthesis pipeline and the Logic-6K dataset, generating texture-consistent yet spatially invalid samples to force the model to learn contextual layout violations instead of texture shortcuts.
\end{itemize}

\section{Related Work}
\label{sec:related work}

Road anomaly segmentation aims to identify unexpected objects in open-world driving scenarios. To prevent models from dangerously misclassifying unknown obstacles as background, current state-of-the-art approaches predominantly rely on segmentation-based uncertainty estimation, moving away from computationally prohibitive reconstruction methods. Despite their progress, these discriminative methods grapple with two pervasive challenges: Softmax-induced overconfidence and a fundamental deficiency in spatial-logical understanding. Existing methods can be categorized into three groups: pixel-wise, object-level and language-level.

\subsection{Pixel-wise Methods via Local Textures}
These methods operate on the premise that texture modes absent from the training distribution constitute anomalies. Extensive research has explored this paradigm: Entropy Maximization~\cite{chan2021entropy} suppresses confidence in anomalous regions; SML~\cite{SML} standardizes pre-trained logits and leverages local visual consistency; PEBAL~\cite{PEBAL} introduces energy-biased abstention learning; and methods like DenseHybrid~\cite{Densehybrid}, RPL~\cite{RPL}, ContMAV~\cite{ContMAV}, and MultiShiftSeg~\cite{MultiShiftSeg} employ logit reinterpretation, residual patterns, or generative augmentations. However, by over-relying on local pixel-level statistics, these methods often neglect the semantic integrity of objects, resulting in fragmented noise. More critically, their lack of long-range dependency modeling severely constrains their ability to identify logical anomalies that violate the overall semantic layout of the scene.

\subsection{Object-level Methods via Transformers and Foundation Models}
To resolve mask fragmentation, these methods leverage attention mechanisms for mask level segmentation. One branch builds upon Mask2Former~\cite{Mask2Former}, utilizing object queries to predict masks directly as Mask2Anomaly~\cite{Mask2Anomaly}, RbA~\cite{RbA}, UNO~\cite{UNO}. Another emerging branch harnesses the zero-shot capabilities of visual foundation models. For instance, S2M~\cite{S2M_CVPR24} converts pixel-level anomaly scores into box prompts to drive SAM\cite{SAM}, while OoDDINO~\cite{OoDDINO} and DetSeg~\cite{DetSeg} employ Grounding DINO~\cite{GroundingDINO} and universal queries to execute a ``detect-all-then-suppress'' paradigm. Although incorporating object-level mechanisms improves structural integrity, these cascaded, multi-stage pipelines inherently incur massive computational and latency overheads, making them highly unfavorable for real-time autonomous driving. Furthermore, their rejection mechanisms remain tied to appearance-based discrimination, overlooking context-dependent anomalies. Meanwhile, S3AD\cite{S3AD} and DDStereo\cite{DDStereo} propose employing stereo disparity guidance for 3D road anomaly detection.

\subsection{Vision-Language Model (VLM) Driven Methods}
Models like CLIP\cite{CLIP} have propelled anomaly detection toward open-vocabulary perception. VL4AD~\cite{VL4AD}, for example, exploits latent cross-modal alignment between image features and textual prompts to identify anomalous patterns. While VLMs successfully harness generalized semantic knowledge, they fundamentally lack an explicit modeling mechanism for spatial-logical relationships. Consequently, they struggle to assess the contextual rationality of an object within a specific layout.

\noindent \textbf{Summary.} Existing paradigms universally suffer from \textit{texture bias}, detecting anomalies via appearance shifts rather than spatial-logic violations. Consequently, they fail on context-dependent anomalies (e.g., a traffic cone is benign on the roadside but hazardous in the lane). To bridge this gap, our SLVS pipeline and cascade-free framework shift the detection paradigm from mere texture novelty to genuine contextual reasoning.

\section{Our Approach}
\label{sec:our approach}

\subsection{General Network Architecture}

\begin{figure}
\centering
\includegraphics[width=\textwidth]{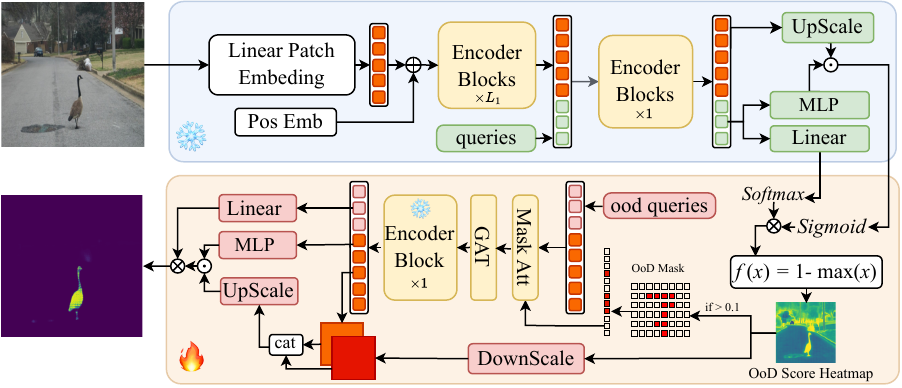}
\caption{Overview of the proposed architecture.}
\label{fig:framework}
\end{figure}

The proposed framework builds upon EoMT~\cite{EoMT}, an Encoder-only Mask Transformer that repurposes a plain Vision Transformer (ViT) architecture for image segmentation without relying on task-specific decoder components. As illustrated in Fig.~\ref{fig:framework}, EoMT concatenates a set of $Q$ learnable semantic queries, denoted as $\mathbf{q} \in \mathbb{R}^{Q \times d}$ (where $d$ is the embedding dimension), to the patch token sequence after the initial $L_1$ encoder blocks. These query and patch tokens are subsequently processed jointly by the remaining $L_2$ ViT blocks via standard multi-head self-attention. Here, the query-to-patch attention implicitly serves the role of cross-attention found in conventional Mask Transformer decoders. Finally, a lightweight mask module---comprising a linear classification head and a three-layer MLP followed by a dot-product with upscaled patch features---is applied to the query outputs to yield semantic mask logits $M \in \mathbb{R}^{Q \times H \times W}$ and class logits $C \in \mathbb{R}^{Q \times K}$. 

On top of this frozen closed-set segmentation foundation, we introduce a lightweight, trainable OoD refinement branch. Instead of cascading a separate computationally heavy model, this branch constructs an independent decoder simply by duplicating the final ViT encoder block. A dedicated set of $Q$ learnable OoD queries $\mathbf{q}_{ood} \in \mathbb{R}^{Q \times d}$ is introduced and concatenated with the patch tokens output by the frozen backbone, followed by joint processing within this independent block. Crucially, the query-to-patch attention in this block is modulated by a binary OoD attention mask derived from the OoD Heatmap (detailed in Sec.~\ref{sec:3.2}). This modulation explicitly restricts the OoD queries to attend exclusively to high-uncertainty spatial regions (Sec.~\ref{sec:3.3}). The updated OoD queries are then decoded by an independent 3-class classification head and a three-layer MLP mask head. The latter is combined with OoD-Heatmap-enriched, upscaled patch features via an Einstein summation (einsum) projection to generate the final anomaly segmentation mask. During training, only the parameters of the OoD branch are updated, while the closed-set segmentation branch remains strictly frozen to preserve its original in-domain segmentation capability and prevent catastrophic forgetting.

\subsection{OoD Heatmap}
\label{sec:3.2}
To effectively localize anomalies, we establish a spatial uncertainty prior by constructing a dense Pixel Out-of-Distribution (OoD) heatmap from the foundational semantic branch. Specifically, given the sigmoid-activated masks $M \in \mathbb{R}^{Q \times (H \cdot W)}$ and softmax-normalized class probabilities $C \in \mathbb{R}^{Q \times K}$ for $K$ known categories ($Q$ is the number of queries), the dense semantic probability map is computed via matrix multiplication as $P = C^\top M$. After reshaping $P$ back to $\mathbb{R}^{K \times H \times W}$, we apply the Maximum Softmax Probability (MSP) formulation to derive the spatial OoD heatmap $S_{ood} \in \mathbb{R}^{H \times W}$:
\begin{align}
    S_{ood}(i, j) = 1 - \max_{k \in \{1, \dots, K\}} P(k, i, j)
\end{align}
The resulting $S_{ood}$ serves a dual purpose: it is processed by a compression block to extract anomaly features that explicitly enrich the latent patch tokens, and it acts as a critical spatial prior to guide the subsequent OoD attention mechanism.

\subsection{OoD Heatmap Guided Spatial Attention}
\label{sec:3.3}

In query-based segmentation backbones, the learned object queries are optimized to explain in-domain classes and can therefore be biased toward dominant background or known-object patterns, making them less sensitive to rare anomalies. To explicitly decouple anomaly representations from the in-domain pathway, we introduce an OoD-Guided Attention refinement in a dedicated OoD branch. This branch uses a spatial prior to restrict the receptive field of OoD queries, forcing them to aggregate features exclusively from highly suspicious regions. In addition, we enhance local spatial-logical reasoning by explicitly modeling neighborhood relations with a lightweight Graph Attention(GAT) module applied to patch tokens before they interact with the OoD queries.

Given the dense OoD heatmap $S_{ood}$ (Sec.~\ref{sec:3.2}), we bilinearly resize it to match the ViT patch grid and binarize it with a fixed threshold $\tau = 0.1$ to obtain a Boolean mask $M \in \{0,1\}^{N_p}$, where $M(j)=1$ indicates a suspicious patch. Let $\mathbf{T} \in \mathbb{R}^{N_p \times d}$ denote the patch tokens and $\mathbf{Q}_{ood} \in \mathbb{R}^{Q \times d}$ denote the learnable OoD queries. We first apply a two-layer GAT over $\mathbf{T}$ using edges constructed from normalized 2D patch distances (via a radius-based graph). Subsequently, we concatenate $\mathbf{Q}_{ood}$ with the updated tokens to form $\mathbf{X} = [\mathbf{Q}_{ood}; \mathbf{T}]$. The OoD refinement is executed by a single duplicated ViT block (rather than a separate heavy decoder) employing masked self-attention. The attention logits are modulated exclusively for query-to-patch connections:

\begin{align}
    \mathrm{Attn}(\mathbf{X}) = \mathrm{Softmax}\!\left(\frac{\mathbf{Q}\mathbf{K}^\top}{\sqrt{d}} + \mathbf{A}\right)\mathbf{V}
\end{align}
where the modulation matrix $\mathbf{A}$ restricts the attention of the OoD queries. Specifically, if $i$ represents the index of an OoD query and $j$ represents the index of a patch token, $\mathbf{A}(i, j)$ is defined as:
\begin{align}
    \mathbf{A}(i, j) = \begin{cases} 
        0, & \text{if } M(j) = 1 \\ 
        -\infty, & \text{if } M(j) = 0 
    \end{cases}
\end{align}
For all other token pairs, $\mathbf{A}(i, j) = 0$. This explicit masking zeroes out the attention weights assigned to non-suspicious patches after the Softmax operation while leaving all other token interactions intact, ensuring the OoD queries focus on anomalous regions. Importantly, the GAT is strictly confined to the OoD refinement branch, leaving the original segmentation pathway entirely unaltered.
\subsection{Training Paradigm and Loss Function}
\label{sec:3.4}

To prevent the model from catastrophically forgetting in-domain (ID) representations while acquiring anomaly detection capabilities, we adopt a Multi-Source Joint Training Paradigm. The model is initialized with weights pre-trained on standard road scene datasets (e.g., Cityscapes\cite{Cityscapes}). During the subsequent anomaly fine-tuning stage, training batches are uniformly sampled from a concatenated dataset containing both ID normal scenes and diverse anomalous scenes (including conventional synthesis datasets and our proposed Logic-6K dataset).

Our anomaly segmentation head is supervised under a standard mask classification paradigm. The overall loss function is formulated as a linear combination of a mask classification loss, a pixel-wise binary cross-entropy (BCE) mask loss, and a Dice loss. Crucially, the incorporation of the Dice loss optimizes the global region-level overlap between the predicted and ground-truth anomaly masks. This formulation is highly effective in mitigating the severe class imbalance inherent in road anomaly detection, as it renders the loss agnostic to the overwhelmingly large number of background pixels, thereby enhancing the model's robustness to anomalous objects of varying scales. The total training loss is defined as:
\begin{align}
    \mathcal{L}_{total} = \lambda_{cls} \mathcal{L}_{cls} + \lambda_{mask} \mathcal{L}_{mask} + \lambda_{dice} \mathcal{L}_{dice}
\end{align}
where $\mathcal{L}_{cls}$, $\mathcal{L}_{mask}$, and $\mathcal{L}_{dice}$ denote the mask classification loss, pixel-wise segmentation loss, and Dice loss, respectively. The hyper-parameters $\lambda_{cls}$, $\lambda_{mask}$, and $\lambda_{dice}$ are the corresponding trade-off weights used to balance these terms.

\subsection{Spatial-Logic Violation Synthesis Pipeline}
\label{sec:3.5}

\begin{figure}[h]
\centering
\includegraphics[width=\textwidth]{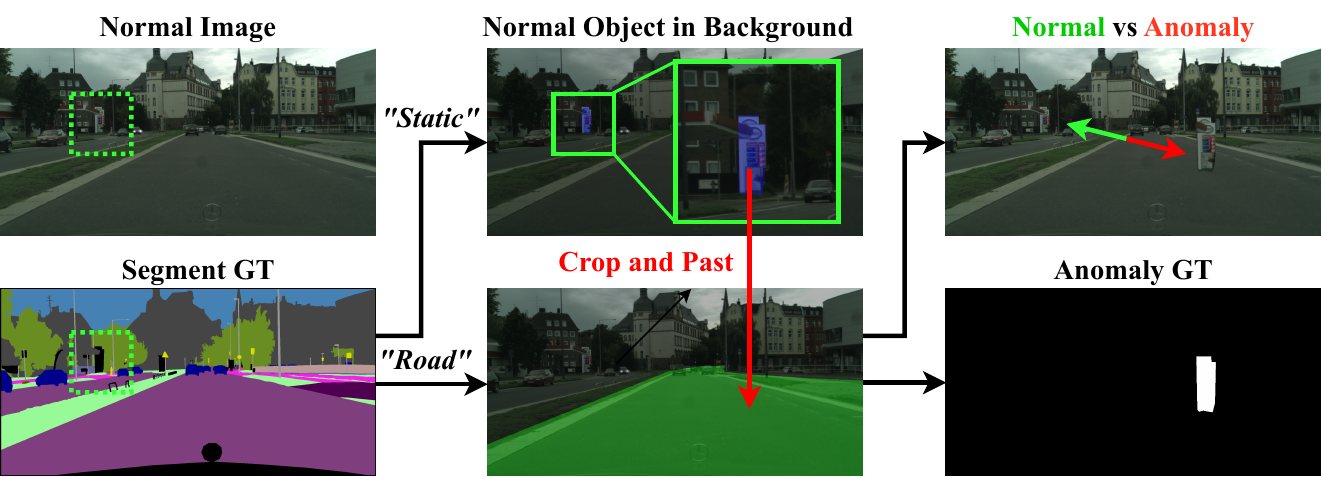}
\caption{Overview of the data generation pipeline}
\label{fig:data_generation}
\end{figure}

Acquiring large-scale, pixel-level annotated anomaly data is challenging due to the long-tail distribution of real-world road obstacles. Existing synthesis paradigms typically paste out-of-domain objects (from COCO) into driving scenes. However, this inadvertently causes models to rely on texture discrepancies rather than comprehending genuine spatial logic. To address this, we introduce the Spatial-Logic Violation Synthesis (SLVS) pipeline (Fig.~\ref{fig:data_generation}). Instead of introducing novel textures, SLVS repurposes indigenous background elements and embeds them onto the road surface. This constructs anomalies that are texture-consistent yet spatially mutually exclusive, forcing the network to abandon texture-based shortcuts and learn contextual violations.

The SLVS pipeline operates in three streamlined stages. \textbf{1) Anomaly Selection:} We establish a Background Object Pool using source-domain annotations (e.g., Cityscapes\cite{Cityscapes}), selecting indigenous objects (e.g., static) that share identical lighting and texture with the normal scene but are strictly prohibited on the road. \textbf{2) Anomaly Synthesis:} Selected instances are randomly embedded into the road area, explicitly disrupting the semantic layout to create a spatial-logic conflict. \textbf{3) Data Cleaning:} A human-in-the-loop filtering process removes physically implausible artifacts (e.g., floating objects, perspective distortions) to guarantee high data fidelity. Following the SLVS protocol, we construct \textbf{Logic-6K}, a synthetic training set built upon Cityscapes\cite{Cityscapes}. After automatic synthesis and manual filtering, Logic-6K provides over 6,000 high-quality images with pixel-level anomaly annotations. Unlike conventional paste-based datasets, Logic-6K preserves visual consistency while explicitly breaking spatial layout priors, making it highly effective for training context-aware anomaly detectors.

\section{Experiments}
\subsection{Experimental Setup}

\noindent \textbf{Metrics.} Performance is measured using Average Precision (AP), Area Under the Receiver Operating Characteristic curve (AuROC), and False Positive Rate at a 95\% True Positive Rate (FPR$_{95}$). AP effectively addresses the inherent class imbalance of anomaly detection, while FPR$_{95}$ serves as a critical safety metric for autonomous driving by quantifying false alarms at high recall levels.

\noindent \textbf{Implementation Details.} We employ a pre-trained ViT-Large (initialized with DINOv2~\cite{dinov2} weights) as the backbone. Input images are resized to $1024 \times 1024$. Standard data augmentations include random color jittering and spatial scaling within $[0.1, 2.0]$. The model is optimized using AdamW with a base learning rate of $1\text{e-}4$, a weight decay of $0.05$.

\subsection{Main Results}

\begin{table}[h]
\centering
\caption{Results on Segment Me If You Can(SMIYC) }
\label{tab:SegmentMeIfYouCan}
\begin{tabular}{l| l |c c |c c}
\toprule
 & & \multicolumn{2}{c|}{Anomaly} & \multicolumn{2}{c}{Obstacle} \\
\cmidrule(lr){3-4} \cmidrule(lr){5-6}
Method & Venue & AP $\uparrow$ & FPR$_{95}$ $\downarrow$ & AP $\uparrow$ & FPR$_{95}$ $\downarrow$ \\
\midrule
PEBAL~\cite{PEBAL} &ECCV'22 &49.10 &40.80&5.00&12.70 \\
RbA~\cite{RbA} & ICCV'23&90.90 &11.60&91.80&0.50  \\
Maskomaly~\cite{maskomaly} & BMVC'23&93.35&6.87&-&- \\
Mask2Anomaly~\cite{Mask2Anomaly} &ICCV'23 &88.70&14.60 &93.30 &0.20    \\
UNO~\cite{UNO} &BMVC'24 &96.33&\textbf{1.98}&93.20&0.20 \\
ContMAV~\cite{ContMAV} &CVPR'24& 90.00 &	2.68 &-&-\\
MultiShiftSeg~\cite{MultiShiftSeg} &NeurIPS'25 &91.92&4.90 & \textbf{95.29}& 0.07 \\
OoDDINO~\cite{OoDDINO} & ACM MM'25&85.60&7.70&94.50&\textbf{0.05} \\
UEM~\cite{UEM} &IJCV'25&95.58&4.70&94.38&0.10\\ 
PixOOD~\cite{pixood} &PAMI'26 & 68.88&	54.33&88.90&0.30    \\ \midrule
LARAD(Ours) & - &\textbf{96.59} & 2.51 &90.61 & 0.07  \\
\bottomrule
\end{tabular}
\end{table}

\noindent \textbf{Results on Online Benchmarks.} Evaluated on the SMIYC\cite{SMIYC} benchmark (Table~\ref{tab:SegmentMeIfYouCan}) against withheld ground truths, our model proves exceptional generalization. In the Anomaly Track, it achieves the highest AP 96.59 and a competitive FPR$_{95}$ (2.51, second only to UNO's 1.98). In the Obstacle Track, it maintains robust performance (90.61 AP, 0.07 FPR$_{95}$). Notably, this outstanding 0.07 FPR$_{95}$ rivals specialized obstacle detectors like MultiShiftSeg~\cite{MultiShiftSeg} and OoDDINO~\cite{OoDDINO}, demonstrating highly reliable handling of small-profile anomalies.

\noindent \textbf{Results on Public Benchmarks.} On RoadAnomaly~\cite{RA} (Table~\ref{tab:performance_comparison_RA}), LARAD achieves state-of-the-art (SOTA) FPR$_{95}$ (3.96) and AuROC (99.14). On Fishyscapes Static~\cite{Fishyscapes} (Table~\ref{tab:performance_comparison_fc lost and found}), we establish a new SOTA across all metrics (FPR$_{95}$ 0.04, AP 97.97, AuROC 99.82). On Fishyscapes Lost \& Found~\cite{Fishyscapes}, we yield 77.81 AP, outperforming RbA (70.81 AP) but trailing PixOOD (93.55 AP). This gap is an expected trade-off: LARAD's reliance on contextual layout inherently limits its sensitivity to the extremely small, isolated obstacles abundant in this subset.

\begin{table}[h]
\centering
\small
\caption{Comparison with state-of-theart methods on Road Anomaly.}
\label{tab:performance_comparison_RA}
\begin{tabular}{l |l |c c c}
\toprule
Method & Venue & FPR$_{95}$$\downarrow$  & AP  $\uparrow$& AuROC $\uparrow$ \\
\midrule
PEBAL~\cite{PEBAL}             & ECCV'22       & 44.58 & 45.10 & 87.63 \\
MGCDA~\cite{MGCDA}                & MM'23         & 42.19 & 50.35 & --    \\
RPL~\cite{RPL}                  & ICCV'23       & 17.74 & 71.60 & 95.72 \\

RbA~\cite{RbA}              & ICCV'23       & 6.92  & 85.42 & 97.99 \\
M2F-EAM~\cite{m2f-eam} & CVPR'23 & 7.70& 69.40 & - \\
Maskomaly~\cite{maskomaly} & BMVC'23&12.00&80.80&-\\
Mask2Anomaly~\cite{Mask2Anomaly} & PAMI'24                     &13.45&79.70&-\\
UNO~\cite{UNO} & BMVC'24&7.40&88.50&-\\
MultiShiftSeg~\cite{MultiShiftSeg}                           & NeurIPS'24     & 7.54 & 90.17 & 97.94\\ 
OoDDINO~\cite{OoDDINO}                             & ACM MM'25        & 4.78  & 87.13 & 98.73 \\
UEM~\cite{UEM} &IJCV'25&8.95&92.86&98.08\\
PixOOD~\cite{pixood} &PAMI'26 & 4.30&\textbf{96.39}&-\\

\hline
LARAD(Ours)&        & \textbf{3.96} & 94.13 &\textbf{99.14} \\
\bottomrule
\end{tabular}
\end{table}

\begin{table}[h!]
\centering
\small
\caption{Results on Fishyscapes Lost \& Found and Static}
\label{tab:performance_comparison_fc lost and found}
\begin{tabular}{l| l| c c c |c c c}
\toprule
 &  & \multicolumn{3}{c|}{FS Lost and Found} & \multicolumn{3}{c}{FS Static} \\
\cmidrule(lr){3-5} \cmidrule(lr){6-8}
Method& Venue &FPR$_{95}$$\downarrow$ & AP $\uparrow$& AuROC $\uparrow$& FPR$_{95}$$\downarrow$ & AP$\uparrow$ & AuROC$\uparrow$ \\
\midrule
SML\cite{SML}&ICCV'21&14.53&36.55&96.88&16.75&48.67&48.65\\

RbA~\cite{RbA} & ICCV'23        & 6.30  & 70.81 &  \textbf{98.62}& - &- &-\\ 
Mask2Anomaly~\cite{Mask2Anomaly} & PAMI'24      & 4.36  & 46.04 &  - & 0.82&95.20 &- \\
UNO~\cite{UNO} & BMVC'24& 2.70&74.80&-&0.30&95.80&-\\
PixOOD~\cite{pixood} &PAMI'26 & \textbf{0.54}&\textbf{93.55}&-&-&-& -\\
\hline
LARAD(Ours) & - & 11.58 & 77.81 & 97.28 & \textbf{0.04} & \textbf{97.97}  & \textbf{99.82}  \\

\bottomrule
\end{tabular}
\end{table}

\subsection{Ablation Studies}

To validate the effectiveness of the proposed components, we conduct extensive ablation experiments on the RoadAnomaly benchmark. As summarized in Table~\ref{tab:ablation studies}, we progressively analyze the contributions of dedicated OoD queries, GAT based spatial attention, and the MSP-based OoD heatmap prior.

\begin{table}[htbp]
\centering
\caption{Ablation on Key Architectural Components on RoadAnomaly}
\label{tab:ablation studies}
\begin{tabular}{c |c | c| c c c}
\toprule
OoD Queries & GAT & Heatmap & AuROC $\uparrow$& AP$\uparrow$ & FPR$_{95}$ $\downarrow$\\
\midrule
\checkmark & \checkmark & MSP  & \textbf{99.14} & \textbf{94.13} & \textbf{3.96} \\
$\times$ & \checkmark  & MSP  & 97.97 & 91.39 & 5.15 \\
\checkmark & $\times$ & MSP  & 98.35 & 92.27 & 6.05 \\
\checkmark & \checkmark & Ones  & 98.65 & 93.32 & 5.57  \\

\bottomrule
\end{tabular}  
\end{table}

\noindent \textbf{Effectiveness of Dedicated OoD Queries:}
Replacing the dedicated OoD queries with shared segmentation queries leads to a performance degradation across all metrics. Specifically, AP drops from 94.13 to 91.39. This indicates that shared queries struggle to disentangle anomaly representations from dominant in-distribution semantics. In contrast, the proposed dedicated OoD queries provide an independent representation space for anomalous regions, enabling more accurate localization of unknown obstacles.

\noindent \textbf{Effectiveness of GAT:}
Removing the proposed spatial attention GAT module causes AP to decrease from 94.13 to 92.27, accompanied by a significant increase of FPR$_{95}$ from 3.96 to 6.05. This degradation demonstrates that unconstrained attention tends to diffuse into dominant background regions, weakening anomaly discrimination. The proposed design effectively suppresses background interference and improves mask precision for spatially inconsistent anomalies.

\noindent \textbf{Effectiveness of the MSP-based OoD Heatmap Prior:}
Replacing the MSP-derived OoD heatmap with a uniform all-one prior also causes noticeable degradation, reducing AP from 94.13 to 93.32 and increasing FPR$_{95}$ from 3.96 to 5.57. This result verifies that the OoD heatmap provides an informative spatial uncertainty prior that guides the refinement branch toward suspicious regions. Without this prior, the OoD branch loses reliable localization cues and becomes more vulnerable to background noise.

Overall, the complete model consistently achieves the best performance across all metrics, demonstrating that the OoD queries, spatially-guided attention mechanism, and uncertainty-aware heatmap prior are highly complementary.

\noindent \textbf{Effect of Training Data Composition:}
Table~\ref{tab:ablation studies data} ablates the training data composition. Using only COCO-Mix yields 98.48 AuROC and 89.38 AP, and adding KITTI-AR improves performance to 98.76 AuROC / 92.08 AP. Incorporating our Logic-6K further boosts results to 99.14 AuROC / 94.13 AP. This progression shows that Logic-6K supplies complementary spatial-layout violation supervision beyond the appearance diversity of COCO-Mix and the realistic road anomalies of KITTI-AR, successfully shifting the model from texture-based discrimination toward contextual spatial reasoning.

\begin{table}[htbp]
\centering
\caption{Ablation on Training Data Composition on RoadAnomaly}
\label{tab:ablation studies data}
\begin{tabular}{c |c |c| c c}
\toprule

COCO-Mix & KITTI-AR & Logic-6k & AuROC & AP\\
\midrule
 \checkmark &   \checkmark & \checkmark & \textbf{99.14} & \textbf{94.13} \\
\checkmark &   \checkmark & $\times$ & 98.76 & 92.08\\
\checkmark &   $\times$ & $\times$ & 98.48 & 89.38\\
\bottomrule
\end{tabular}
\end{table}

\subsection{Efficiency Analysis}
Table~\ref{tab:efficiency} compares inference efficiency. Acknowledging hardware discrepancies across baselines (e.g., RTX A100/A5000), this serves as an indicative reference. Evaluated on a single RTX 4090, LARAD achieves 13.76 FPS (0.0726 s/frame), significantly outperforming RPL~\cite{RPL}, RbA~\cite{RbA}, S2M~\cite{S2M_CVPR24}, and SOTA-RBA\cite{SOTA}. This speed advantage stems from our cascade-free design: unlike latency-heavy multi-stage pipelines as S2M~\cite{S2M_CVPR24} that invoke separate foundation models, LARAD localizes anomalies entirely within a single compact network, yielding a superior accuracy-efficiency trade-off.

\begin{table}
\centering
\caption{Inference efficiency comparison. FPS is computed from reported per-frame latency. Results span different hardware platforms and serve as an indicative baseline.}
\label{tab:efficiency}
\begin{tabular}{llcc}
\toprule
\textbf{Method} & \textbf{Hardware} & \textbf{Latency (s)} $\downarrow$ & \textbf{FPS} $\uparrow$ \\
\midrule
RPL~\cite{RPL} & RTX A5000 & 0.2170 & 4.61 \\
RbA~\cite{RbA} & RTX A100 & 0.1833 & 5.46 \\
S2M~\cite{S2M_CVPR24} & RTX A5000 & 0.2767 & 3.61 \\
SOTA-RBA~\cite{SOTA} & RTX A100 & 0.3667 & 2.73 \\
\midrule
LARAD(Ours) & RTX 4090 & \textbf{0.0726} & \textbf{13.76} \\
\bottomrule
\end{tabular}
\end{table}

\section{Visualization}
\begin{figure}
\centering
\includegraphics[width=0.5\textwidth]{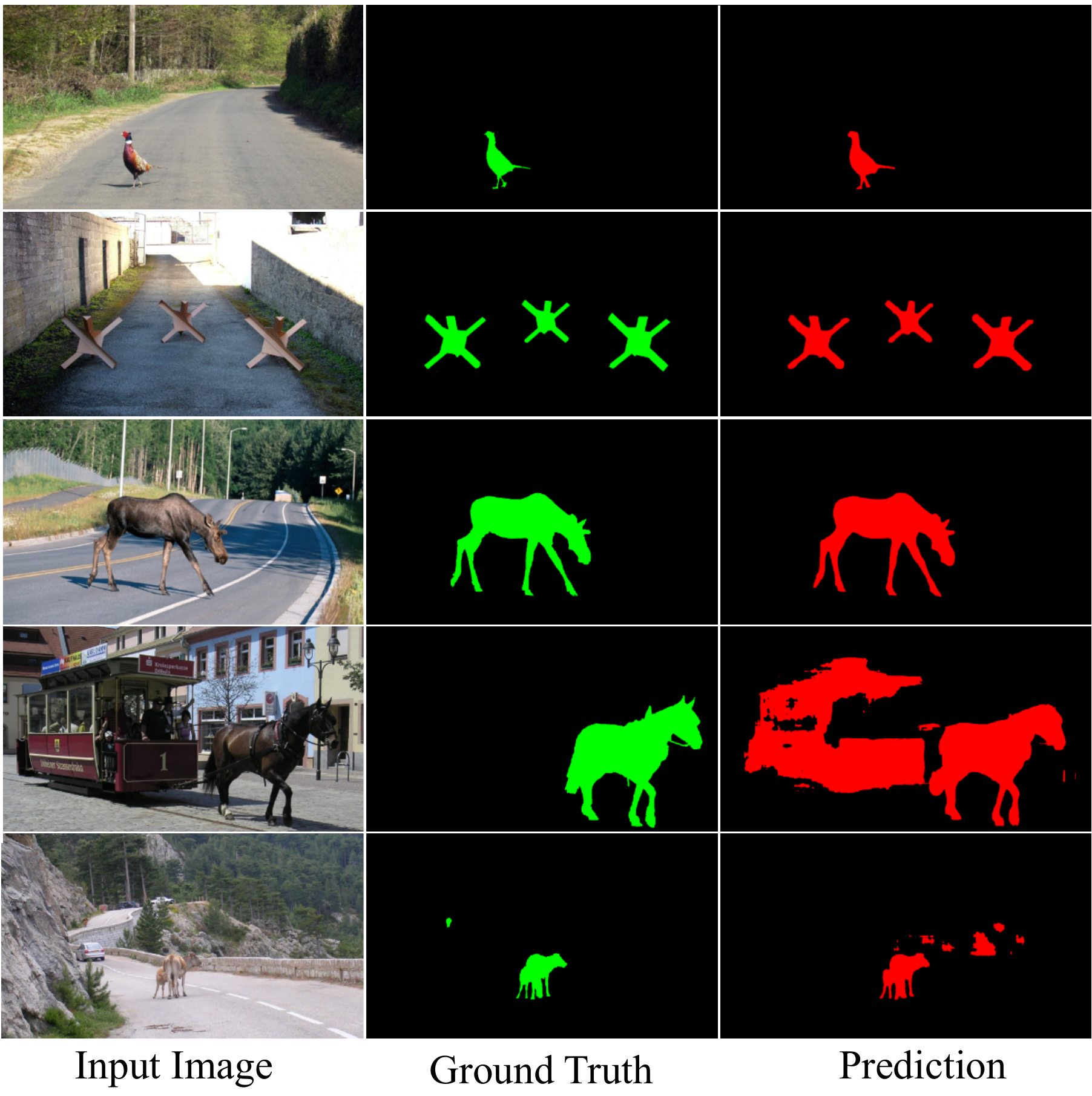}
\caption{Visualization of successful and failure cases of road anomaly segmentation.}
\label{fig:result}
\end{figure}

Figure~\ref{fig:result} shows success and failure cases of our algorithm. Rows 1–3 present anomalous objects on the road, where details and edges are detected well. Rows 4–5 illustrate failure cases: a carriage and distant mountains are detected as false positives. This occurs because their texture and layout logic did not appear in the training scenes and are thus judged as novel anomalous texture and layout logic, indicating that our algorithm still carries a risk of false positives.

\section{Conclusion}

We propose LARAD, a layout-aware road anomaly detection framework that overcomes the latency and appearance-bias limitations of cascaded paradigms. By augmenting a standard segmentation model with a lightweight, OoD-guided attention branch, LARAD achieves precise anomaly localization while preserving closed-set accuracy and single-model efficiency. To eliminate texture-novelty reliance, we introduce the Spatial-Logic Violation Synthesis (SLVS) pipeline and the Logic-6K dataset. By generating texture-consistent yet spatially invalid samples, we compel the network to learn genuine contextual layout violations. Extensive benchmarks (RoadAnomaly, Fishyscapes, SMIYC) confirm LARAD establishes a new state-of-the-art accuracy-efficiency trade-off. Future work targets real-time edge deployment on Jetson Orin via TensorRT and quantization.

\section*{Acknowledgements}
This work was supported in part by the 6G Science and Technology Innovation and Future Industry Cultivation Special Project of Shanghai Municipal Science and Technology Commission under Grant 24DP1501001, in part by the National High Quality Program under Grant TC220H07D, and in part by the Xi'an Jiaotong-Liverpool University under Grant for ILAI.
\bibliographystyle{splncs04}
\bibliography{mybibliography}

@String(CVPR  = {IEEE Conf. Comput. Vis. Pattern Recog.})

@String(ICCV  = {Int. Conf. Comput. Vis.})

@String(BMVC  = {Brit. Mach. Vis. Conf.})

@String(CVPR  = {CVPR})

@String(ICCV  = {ICCV})

@String(BMVC  =	{BMVC})

@inproceedings{Densehybrid,
  title={Densehybrid: Hybrid anomaly detection for dense open-set recognition},
  author={Grci{\'c}, Matej and Bevandi{\'c}, Petra and {\v{S}}egvi{\'c}, Sini{\v{s}}a},
  booktitle={European Conference on Computer Vision},
  pages={500--517},
  year={2022},
  organization={Springer}
}

@inproceedings{PEBAL,
  title={Pixel-wise energy-biased abstention learning for anomaly segmentation on complex urban driving scenes},
  author={Tian, Yu and Liu, Yuyuan and Pang, Guansong and Liu, Fengbei and Chen, Yuanhong and Carneiro, Gustavo},
  booktitle={European Conference on Computer Vision},
  pages={246--263},
  year={2022},
  organization={Springer}
}

@InProceedings{RPL,
    author    = {Liu, Yuyuan and Ding, Choubo and Tian, Yu and Pang, Guansong and Belagiannis, Vasileios and Reid, Ian and Carneiro, Gustavo},
    title     = {Residual Pattern Learning for Pixel-Wise Out-of-Distribution Detection in Semantic Segmentation},
    booktitle = {Proceedings of the IEEE/CVF International Conference on Computer Vision (ICCV)},
    month     = {October},
    year      = {2023},
    pages     = {1151-1161}
}

@InProceedings{S2M_CVPR24,
    author    = {Zhao, Wenjie and Li, Jia and Dong, Xin and Xiang, Yu and Guo, Yunhui},
    title     = {Segment Every Out-of-Distribution Object},
    booktitle = {Proceedings of the IEEE/CVF Conference on Computer Vision and Pattern Recognition (CVPR)},
    month     = {June},
    year      = {2024},
    pages     = {3910-3920}
}

@inproceedings{chan2021entropy,
  title={Entropy maximization and meta classification for out-of-distribution detection in semantic segmentation},
  author={Chan, Robin and Rottmann, Matthias and Gottschalk, Hanno},
  booktitle={Proceedings of the ieee/cvf international conference on computer vision},
  pages={5128--5137},
  year={2021}
}

@inproceedings{SML,
    author    = {Jung, Sanghun and Lee, Jungsoo and Gwak, Daehoon and Choi, Sungha and Choo, Jaegul},
    title     = {Standardized Max Logits: A Simple yet Effective Approach for Identifying Unexpected Road Obstacles in Urban-Scene Segmentation},
    booktitle = {Proceedings of the IEEE/CVF International Conference on Computer Vision (ICCV)},
    month     = {October},
    year      = {2021},
    pages     = {15425-15434}
}

@inproceedings{ContMAV,
    author    = {Sodano, Matteo and Magistri, Federico and Nunes, Lucas and Behley, Jens and Stachniss, Cyrill},
    title     = {Open-World Semantic Segmentation Including Class Similarity},
    booktitle = {Proceedings of the IEEE/CVF Conference on Computer Vision and Pattern Recognition (CVPR)},
    month     = {June},
    year      = {2024},
    pages     = {3184-3194}
}

@article{MultiShiftSeg,
  title={Generalize or detect? towards robust semantic segmentation under multiple distribution shifts},
  author={Gao, Zhitong and Li, Bingnan and Salzmann, Mathieu and He, Xuming},
  journal={Advances in Neural Information Processing Systems},
  volume={37},
  pages={52014--52039},
  year={2024}
}

@ARTICLE{Mask2Anomaly,
  author={Rai, Shyam Nandan and Cermelli, Fabio and Caputo, Barbara and Masone, Carlo},
  journal={IEEE Transactions on Pattern Analysis and Machine Intelligence}, 
  title={Mask2Anomaly: Mask Transformer for Universal Open-Set Segmentation}, 
  year={2024},
  volume={46},
  number={12},
  pages={9286-9302}}

@inproceedings{SOTA,
  title={Segmenting objectiveness and task-awareness unknown region for autonomous driving},
  author={Zheng, Mi and Yang, Guanglei and Huang, Zitong and Guo, Zhenhua and Han, Kevin and Zuo, Wangmeng},
  booktitle={Proceedings of the 33rd ACM International Conference on Multimedia},
  pages={1433--1442},
  year={2025}
}

@article{VL4AD,
  title={VL4AD: Vision-Language Models Improve Pixel-wise Anomaly Detection},
  author={Zhong, Liangyu and Sicking, Joachim and H{\"u}ger, Fabian and Gottschalk, Hanno},
  journal={arXiv preprint arXiv:2409.17330},
  year={2024}
}

@inproceedings{UNO,
  title={Outlier detection by ensembling uncertainty with negative objectness},
  author={Deli{\'c}, Anja and Grci{\'c}, Matej and {\v{S}}egvi{\'c}, Sini{\v{s}}a},
  booktitle={The 35th British Machine Vision Conference},
  pages={1--27},
  year={2024},
  organization={Glasgow: British Machine Vision Association}
}

@inproceedings{RbA,
    author    = {Nayal, Nazir and Yavuz, Misra and Henriques, Jo\~ao F. and G\"uney, Fatma},
    title     = {RbA: Segmenting Unknown Regions Rejected by All},
    booktitle = {Proceedings of the IEEE/CVF International Conference on Computer Vision (ICCV)},
    month     = {October},
    year      = {2023},
    pages     = {711-722}
}

@inproceedings{OoDDINO,
  title={OoDDINO: A Multi-level Framework for Anomaly Segmentation on Complex Road Scenes},
  author={Liu, Yuxing and Zhang, Ji and Zhou, Xuchuan and Xiao, Jingzhong and Yang, Huimin and Zhong, Jiaxin},
  booktitle={Proceedings of the 33rd ACM International Conference on Multimedia},
  pages={2673--2682},
  year={2025}
}

@inproceedings{DetSeg,
  author    = {Zhu, Huachao and Liu, Zelong and Sun, Zhichao and Zou, Yuda and Xia, Gui-Song and Xu, Yongchao},
    title     = {Beyond Pixel Uncertainty: Bounding the OoD Objects in Road Scenes},
    booktitle = {Proceedings of the IEEE/CVF International Conference on Computer Vision (ICCV)},
    month     = {October},
    year      = {2025},
    pages     = {8472-8481}
}

@InProceedings{CLIP,
  title = 	 {Learning Transferable Visual Models From Natural Language Supervision},
  author =       {Radford, Alec and Kim, Jong Wook and Hallacy, Chris and Ramesh, Aditya and Goh, Gabriel and Agarwal, Sandhini and Sastry, Girish and Askell, Amanda and Mishkin, Pamela and Clark, Jack and Krueger, Gretchen and Sutskever, Ilya},
  booktitle = 	 {Proceedings of the 38th International Conference on Machine Learning},
  pages = 	 {8748--8763},
  year = 	 {2021},
  volume = 	 {139},
  month = 	 {18--24 Jul},
  publisher =    {PMLR},
  
}

@InProceedings{EoMT,
    author    = {Kerssies, Tommie and Cavagnero, Niccol\`o and Hermans, Alexander and Norouzi, Narges and Averta, Giuseppe and Leibe, Bastian and Dubbelman, Gijs and de Geus, Daan},
    title     = {Your ViT is Secretly an Image Segmentation Model},
    booktitle = {Proceedings of the IEEE/CVF Conference on Computer Vision and Pattern Recognition (CVPR)},
    month     = {June},
    year      = {2025},
    pages     = {25303-25313}
}

@inproceedings{GroundingDINO,
  title={Grounding dino: Marrying dino with grounded pre-training for open-set object detection},
  author={Liu, Shilong and Zeng, Zhaoyang and Ren, Tianhe and Li, Feng and Zhang, Hao and Yang, Jie and Jiang, Qing and Li, Chunyuan and Yang, Jianwei and Su, Hang and others},
  booktitle={European conference on computer vision},
  pages={38--55},
  year={2024},
  organization={Springer}
}

@inproceedings{SMIYC,
 author = {Chan, Robin and Lis, Krzysztof and Uhlemeyer, Svenja and Blum, Hermann and Honari, Sina and Siegwart, Roland and Fua, Pascal and Salzmann, Mathieu and Rottmann, Matthias},
 booktitle = {Proceedings of the Neural Information Processing Systems Track on Datasets and Benchmarks},
 editor = {J. Vanschoren and S. Yeung},
 pages = {},
 title = {SegmentMeIfYouCan: A Benchmark for Anomaly Segmentation},
 volume = {1},
 year = {2021}
}

@InProceedings{RA,
author = {Lis, Krzysztof and Nakka, Krishna and Fua, Pascal and Salzmann, Mathieu},
title = {Detecting the Unexpected via Image Resynthesis},
booktitle = {Proceedings of the IEEE/CVF International Conference on Computer Vision (ICCV)},
month = {October},
pages={2152--2161},
year = {2019}
}

@article{Fishyscapes,
  title={The fishyscapes benchmark: Measuring blind spots in semantic segmentation},
  author={Blum, Hermann and Sarlin, Paul-Edouard and Nieto, Juan and Siegwart, Roland and Cadena, Cesar},
  journal={International Journal of Computer Vision},
  volume={129},
  number={11},
  pages={3119--3135},
  year={2021},
  publisher={Springer}
}

@article{
dinov2,
title={{DINO}v2: Learning Robust Visual Features without Supervision},
author={Maxime Oquab and Timoth{\'e}e Darcet and Th{\'e}o Moutakanni and Huy V. Vo and Marc Szafraniec and Vasil Khalidov and Pierre Fernandez and Daniel HAZIZA and Francisco Massa and Alaaeldin El-Nouby and others},
journal={Transactions on Machine Learning Research},
issn={2835-8856},
year={2024}
}

@inproceedings{maskomaly,
  title={Maskomaly: Zero-Shot Mask Anomaly Segmentation},
  author={Ackermann, Jan and Sakaridis, Christos and Yu, Fisher},
  booktitle={The British Machine Vision Conference (BMVC)},
  year={2023}
}

@article{pixood,
  title={Pixood: Pixel-level out-of-distribution detection},
  author={Voj{\'\i}{\v{r}}, Tom{\'a}{\v{s}} and Jan, {\v{S}}ochman and Matas, Ji{\v{r}}{\'\i}},
  journal={IEEE Transactions on Pattern Analysis and Machine Intelligence},
  year={2026},
  pages={1-13},
  publisher={IEEE}
}

@article{UEM,
  title={A likelihood ratio-based approach to segmenting unknown objects},
  author={Nayal, Nazir and Shoeb, Youssef and G{\"u}ney, Fatma},
  journal={International Journal of Computer Vision},
  volume={133},
  number={10},
  pages={6860--6872},
  year={2025},
  publisher={Springer}
}

@inproceedings{MGCDA,
  title={Improving anomaly segmentation with multi-granularity cross-domain alignment},
  author={Zhang, Ji and Wu, Xiao and Cheng, Zhi-Qi and He, Qi and Li, Wei},
  booktitle={Proceedings of the 31st ACM international conference on multimedia},
  pages={8515--8524},
  year={2023}
}

@inproceedings{SAM,
  title={Segment anything},
  author={Kirillov, Alexander and Mintun, Eric and Ravi, Nikhila and Mao, Hanzi and Rolland, Chloe and Gustafson, Laura and Xiao, Tete and Whitehead, Spencer and Berg, Alexander C and Lo, Wan-Yen and others},
  booktitle={Proceedings of the IEEE/CVF international conference on computer vision},
  pages={4015--4026},
  year={2023}
}

@inproceedings{Mask2Former,
  title={Masked-attention mask transformer for universal image segmentation},
  author={Cheng, Bowen and Misra, Ishan and Schwing, Alexander G and Kirillov, Alexander and Girdhar, Rohit},
  booktitle={Proceedings of the IEEE/CVF conference on computer vision and pattern recognition},
  pages={1290--1299},
  year={2022}
}

@inproceedings{m2f-eam,
    author    = {Grci\'c, Matej and \v{S}ari\'c, Josip and \v{S}egvi\'c, Sini\v{s}a},
    title     = {On Advantages of Mask-Level Recognition for Outlier-Aware Segmentation},
    booktitle = {Proceedings of the IEEE/CVF Conference on Computer Vision and Pattern Recognition (CVPR) Workshops},
    month     = {June},
    year      = {2023},
    pages     = {2937-2947}

}

@inproceedings{Cityscapes,
author = {Cordts, Marius and Omran, Mohamed and Ramos, Sebastian and Rehfeld, Timo and Enzweiler, Markus and Benenson, Rodrigo and Franke, Uwe and Roth, Stefan and Schiele, Bernt},
title = {The Cityscapes Dataset for Semantic Urban Scene Understanding},
booktitle = {Proceedings of the IEEE Conference on Computer Vision and Pattern Recognition (CVPR)},
month = {June},
year = {2016}
}

@ARTICLE{S3AD,
  author={Mu, Shiyi and Gu, Zichong and Lyu, Hanqi and Gao, Yilin and Xu, Shugong},
  journal={IEEE Internet of Things Journal}, 
  title={Stereo-Based 3-D Anomaly Object Detection for Autonomous Driving: A New Dataset and Baseline}, 
  year={2026},
  volume={13},
  number={14},
  pages={31138-31149}}

@article{DDStereo,
  title={DDStereo: Efficient Dual Decoder Transformers for Stereo 3D Road Anomaly Detection},
  author={Mu, Shiyi and Gu, Zichong and Ai, Zhiqi and Gao, Yilin and Xu, Shugong},
  journal={arXiv preprint arXiv:2606.24805},
  year={2026}
}
\end{document}